\documentclass{article}

\usepackage{microtype}
\usepackage{graphicx}
\usepackage{subfigure}
\usepackage{textgreek}
\usepackage{booktabs} % for professional tables

\usepackage{hyperref}

 \usepackage[accepted]{tccmlicml2021}

\icmltitlerunning{Quantification of Carbon Sequestration in Urban Forests}% using Deep Learning}

\begin{document}

\twocolumn[
\icmltitle{Quantification of Carbon Sequestration in Urban Forests}% using Deep Learning}




\begin{icmlauthorlist}
\icmlauthor{Levente J Klein}{ibm}
\icmlauthor{Wang Zhou}{ibm}
\icmlauthor{Conrad M Albrecht}{ibm,dlr}
\end{icmlauthorlist}

\icmlaffiliation{ibm}{IBM Research, Yorktown Heights, NY, USA}
\icmlaffiliation{dlr}{German Aerospace Center (DLR), We\ss ling, Germany}

\icmlcorrespondingauthor{Levente Klein}{kleinl@us.ibm.com}
\icmlcorrespondingauthor{Wang Zhou}{wang.zhou@ibm.com}

\icmlkeywords{Deep Learning, Carbon sequestration, tree species}

\vskip 0.3in
]



\printAffiliationsAndNotice{}  % leave blank if no need to mention equal contribution

\begin{abstract}
Vegetation, trees in particular, sequester carbon by absorbing carbon dioxide from the atmosphere. However, the lack of efficient quantification methods of carbon stored in trees renders it difficult to track the process. We present an approach to estimate the carbon storage in trees based on fusing multi-spectral aerial imagery and LiDAR data to identify tree coverage, geometric shape, and tree species---key attributes to carbon storage quantification.
We demonstrate that tree species information and their three-dimensional geometric shapes can be estimated from aerial imagery in order to determine the tree's biomass. Specifically, we estimate a total of $52,000$ tons of carbon sequestered in trees for New York City's borough Manhattan.
% Trees can sequester carbon and offset Carbon Dioxide from atmosphere. Periodic acquisition of multispectral aerial imagery combined with machine learning, can automatically detect individual trees, track tree's growth rates, and classify trees based on their species characteristics. Using Random Forest and ResNet34 it is demonstrated that individual trees can be detected in the imagery and trees species information and their three dimensional geometric sizes can be estimated. From tree species and their geometric sizes, a high resolution map of carbon stored in Manhattan, New York City trees estimates a total carbon storage of 52000 tons in all trees.
\end{abstract}
\section{Introduction}

Recent environmental reports underline the pressing need for the elimination of Green House Gases (GHG) from the atmosphere in order to bring the carbon dioxide level to the pre-industrial norm \cite{ipcc}. Carbon removing techniques span from scrubbing emission sources, manufacturing carbon trapping materials, and sequestering carbon in trees or soil. One popular idea proposed recently is the afforestation of 900 billion hectares of land \cite{Bastin19}, which has the potential to offset more than 200 megatons of carbon from the atmosphere. In the emerging carbon trading market, companies may purchase forested land to offset their GHG emission and reduce carbon footprints\cite{lujtens19}. There is a need for tools and platforms able to quantify in near real time and track GHG emissions and total carbon offsets. Such tools may need to estimate the total carbon stored in trees or in soil multiple times a year to support a fair and transparent carbon trading market.

Currently, carbon sequestration is estimated by a plethora of proprietary tools and models, making it hard to compare side by side carbon sequestration models. Carbon storage estimates typically rely on generic models where shape, density, and species distribution of trees is surveyed from small sample plots. Subsequently, figures on larger geographies and environments get extrapolated \cite{Sileshi14}.\\
%\textcolor{red}{[any related references?]}
%tree sizes, trees density, and forest distribution are interpolated from a small subset of ground truth data and translated to a different geographies or environment. 
% It is recognized that carbon capture potential of trees are impacted by the species, age, growth rate, and habitat environment of trees. 
The maximum amount of carbon captured by a tree is predominantly limited by its geometric size which, in turn, is bounded by physics such as water transport from roots to leaves \cite{Koch04}. Hence, knowledge of tree coverage, their geometric sizes and species characteristics is crucial in providing accurate carbon storage estimates. At the same time the task is technically challenging since such information is not readily available for the majority of locations on the planet \cite{Chave14}.
% Furthermore, estimates of tree scaling relationship are based on allometric estimates that changes significantly across the globe\cite{Chave05}. 

In this work, we propose to exploit remote sensing data in order to determine tree coverage, estimate tree's geometric shapes \& species to ultimately quantify the carbon sequestration in those trees at scale. 
Specifically, we train machine learning models to analyze aerial imagery to determine tree coverage, to classify their species, and to determine the local allometric relation for each tree species. LiDAR data from a sample region is utilized to locally calibrate the biomass estimation model. %Once carbon capture model is calibrated, 
Model inference does not require LiDAR data. Hence, the learnt relation may get exploited in regions with only aerial imagery available. 
%Trees distribution, geometric sizes and tree species information are used to quantify the total carbon stored in each tree, providing 
As an illustration, we generate a high-resolution map of carbon sequestration in New York City's urban forest.

\begin{figure*}[ht]
% \vskip 0.2in
\begin{center}
\centerline{\includegraphics[width=1.6\columnwidth]{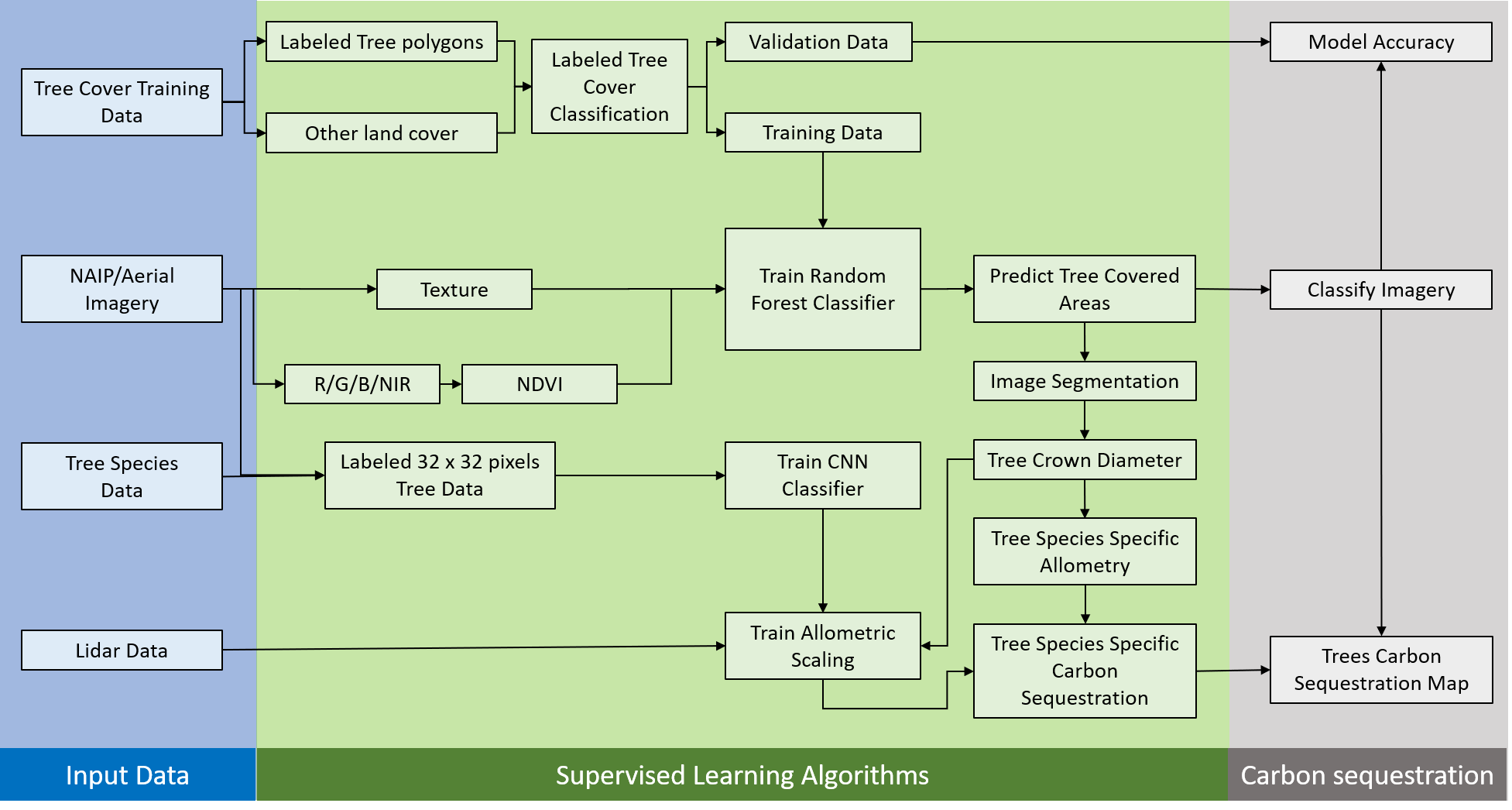}}
\caption{Data processing workflow to quantify carbon sequestration in trees based on aerial images, LiDAR data, and tree species data.}
\label{fig:pipeline}
\end{center}
% \vskip -0.2in
\end{figure*}

\section{Related work}
\label{sec:related}

\paragraph{Tree allometric estimation.} The allometric scaling relates tree height to tree crown diameter, and tree dimensions to tree biomass. Such correlations are common practice in forestry research \cite{Chave05}. Estimates of tree dimensions is important to infer the total biomass which, in turn, relates to the carbon stored in trees.  The literature poses the quest to establish a generic scaling law correlating tree height with tree lateral dimensions \cite{klein19} to better estimate the above ground biomass \cite{Chave14}. Since the scaling relation strongly depends on tree species and tree age in addition to the locations trees grow in (urban vs.\ rural areas), carbon estimate models need local calibration based on tree species and geographical conditions.

\paragraph{Tree species classification.} Quantification of carbon sequestered in trees is limited by the detailed knowledge of the tree's geometrical dimensions and their species. Current advances in image processing enable plot-level tree classification including estimates of tree size \cite{Guan15}. We propose to generalize the image-based estimation method to identify individual trees, and to classify their species \cite{zhou20} for urban forests exploiting \emph{only} aerial imagery accompanied by labeled tree species data. 

\paragraph{Tree total biomass.} Estimates of carbon captured in trees are based on knowing the tree's volume and its wood density which, in turn, strongly depends on the tree species. In order to determine the volume, the tree crown diameter and the tree height need to be determined along with the tree species. The above ground biomass (AGB) may get approximated by \cite{Chave05}:
\begin{equation}
\label{eq:agb}
\mathrm{AGB} = F \times \rho \times (\frac{\pi D ^{2}}{4}) \times H,
\end{equation}
where $H$ represents tree height, $D$ denotes tree canopy diameter, $\rho$ specifies the tree dry mass density, and $F$ is a form factor that takes into account the tree's shape. The shape factor $F$ typically varies in $0.01$ to $1$ depending on the trunk's shape \cite{Chave14}.
%\textcolor{red}{[}It is estimated that around 30\% of the carbon is stored in the roots \cite{Cairns97}. 

The below ground biomass (BGB) is assumed as $0.3 \times \mathrm{AGB}$ \cite{Cairns97} and the total biomass of a tree is specified by the sum of AGB and BGB, i.e. $1.3 \times \mathrm{AGB}$.

\section{Method}

Our pipeline for quantification of carbon sequestration employ aerial imagery and LiDAR data. Fig.~\ref{fig:pipeline} illustrates the data processing steps and machine learning models applied in order to determine tree coverage, tree geometric sizes, and tree species for carbon sequestration estimates. The main machine learning components are: (1) a Random Forest Classifier to identify tree-covered areas combined with image segmentation to delineate individual trees, (2) Deep Learning models to classify tree species, and (3) a carbon calculator tool to estimate the total biomass and carbon sequestered in trees. 
% The carbon sequestration estimates are built on open source data, enabling government agencies and the public to run carbon estimate models. The data analyzed are freely available from government agencies and tree species data used in this study is open data available from NYC \cite{NYC15}. 

\subsection{Data}

\paragraph{2D Imagery.} The National Agriculture Imagery Program (NAIP) acquires aerial imagery every other year at a spatial resolution of 0.6 m on U.S.\ national scale. Multi-spectral bands of red, green, blue and near-infrared are simultaneously collected during full leaves season. NAIP has been consistently collected for the past two decades, making the data source an excellent candidate to track tree coverage, tree growths, and to detect changes in land coverage.
 
\paragraph{3D LiDAR.} LiDAR 3D point clouds are used to extract tree heights, and to calibrate the allometric relations for each tree species. Compared to aerial imagery, LiDAR data is much more expensive to collect, thus unavailable for a major fraction of geographies. Therefore, it is critical to estimate tree height from 2D imagery in order to scale to large geographies where only aerial imagery is readily available.

\paragraph{Tree Species.} Labeled tree species data is available from many municipalities \cite{NYC15} as part of their effort to quantify the benefit of urban forests \cite{opentree}. Typically, data collection is crowd-sourced with specific tree attributes getting captured such as: tree species, tree location, and diameter at breast height. 
%Since such data doesn't exist for many parts of the world, we take the tree  inventory data for NYC and we  generate training data for Deep Learning (DL) models and demonstrate how our method may work using only dominant tree species. We note that this approach may introduce uncertainty is the carbon estimates for trees as the overall diversity in tree species is reduced. Since only the 4 most dominant tree species data is used to train the DL model, current model needs can be expanded to more species but the tree species used in this study represent  60\% of all labeled trees. 

\subsection{Tree detection and segmentation}

We utilize spectral information from remote sensing images to delineate trees from other land cover classes. Employing the red and near-infrared spectral information, we compute the Normalized Difference Vegetation Index (NDVI) \cite{ndvi} from the NAIP imagery. The NDVI is commonly used to separate vegetation from other classes like roads, buildings, bare land, and/or water. Within the vegetation class, sub-classification is achieved by training a Random Forest (RF) model to distinguish trees from grasses or bushes after incorporating additional information like image texture. 
%Such information is quantified within a sliding window of $5\times5$ pixels where the maximum, minimum, mean, average, and standard deviation gets calculated for each or a combination of spectral bands. The spectral and textural input is consumed by the RF classifier to identify areas covered by trees. 

Once the tree mask is generated, %random points are generated to fall within the regions that are covered by trees and a growing region 
segmentation algorithms such as ``watershed'' \cite{watershed} is applied to cluster pixels that share common spectral and textural characteristics. The clustered pixels is converted to (vector) polygons for identification of the tree crown boundary in order to determine the canopy diameter. Tree crown diameters are then used to correlate against tree heights to separately establish the allometric equation for each tree species.

\begin{figure}[t]
% \vskip 0.2in
\begin{center}
\centerline{\includegraphics[width=\columnwidth]{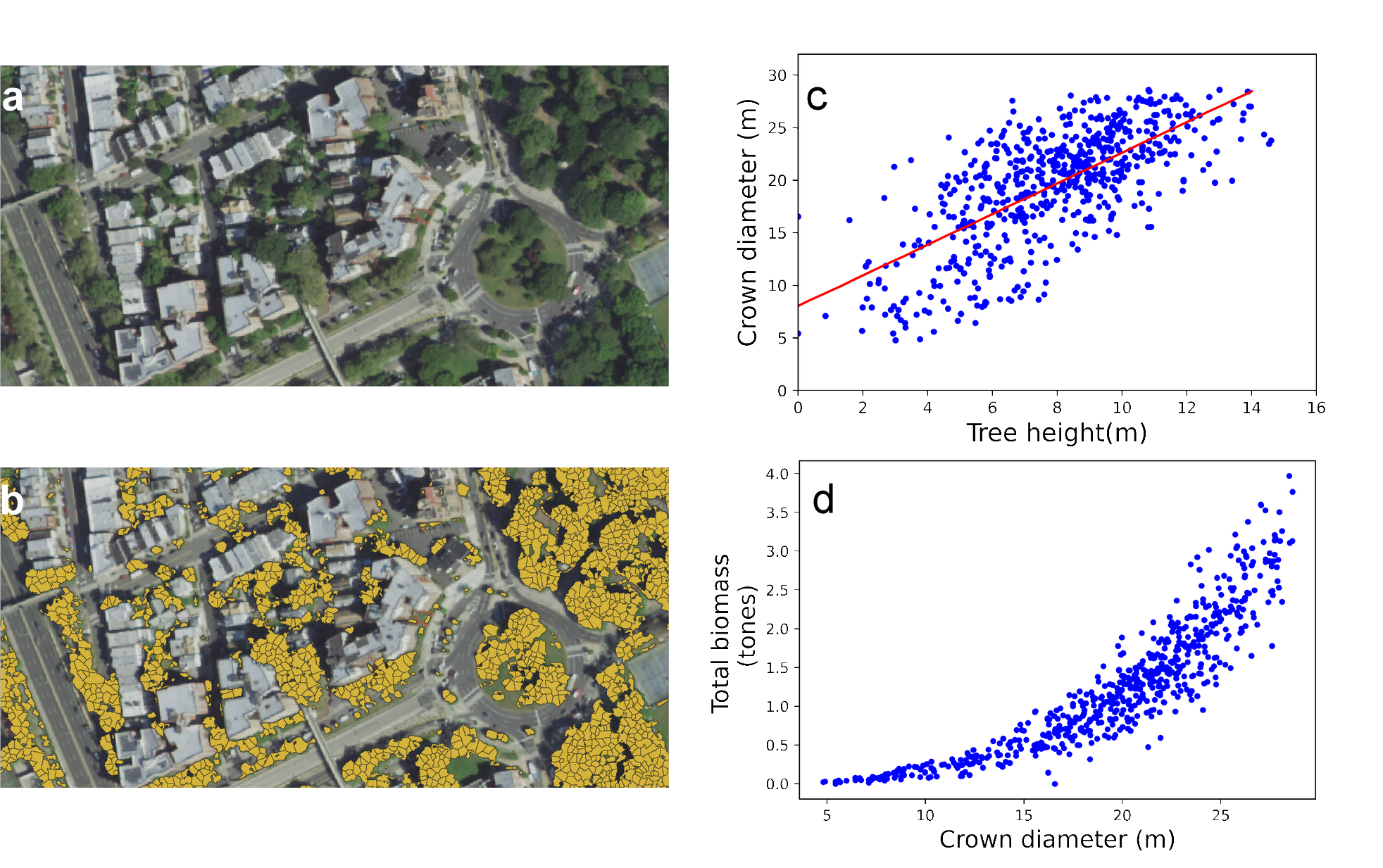}}
\caption{Multi-spectral NAIP imagery (a) and its corresponding segmentation for individual tree crowns (b). The allometric relation between tree crown diameter and tree height for Pin oak tree is illustrated (c) alongside the estimates of its corresponding total biomass (d).}
\label{fig:fitting}
\end{center}
% \vskip -0.2in
\end{figure}
 
\subsection{Tree height estimation}

By virtue of a tree species-specific allometric equation, the height of a tree may get estimated from its crown diameters. In our work, the allometric equation is modeled by a linear fit\footnote{In some cases a more complex relationship may exist between tree crown diameter and tree height \cite{Chave05}. Accordingly, the linear approximation may over or under represent biomass estimates.} mapping the crown size extracted from NAIP imagery to the tree height ground truth extracted from LiDAR data. Once established, the model is applied to areas where no LiDAR data is available.

\subsection{Tree species classification}

%Deep neural networks is the primary method to classify images with the help of large labeled datasets like  ImageNet \cite{imagenet}. The neural networks are trained based on labeled training images. 
Since tree species information is vital to estimate carbon storage, we train a convolutional neural network (CNN) to classify tree species from NAIP imagery.
In our approach, the NAIP images get diced into $32\times32$ tiles. In contrast to standard models of RGB channels, we harness all four spectral bands of the NAIP imagery. The neural network represents a modified ResNet34 \cite{resnet} which allows four-channel images as input. The training data is pre-processed by cropping the NAIP data centered around each location of the labeled trees. The trained model is run across all target areas to generate corresponding tree species maps.

% The networks is customized by allowing \emph{flexible} input formats so that the models can be applied to arbitrary number of data layers. This design is crucial as it enables training network with complex data input combination like spectral bands from NAIP or bands from Lidar data. As an example, here we apply a modified \emph{ResNet34} \cite{resnet} that is trained using as input 4 NAIP bands (Fig 2). 

% The ResNet34 model achieves an accuracy of $80\%$, for the four tree species classes (London planetree, Honeylocust, Callery Pear and Pin Oak). The fifth class was collecting all the others pixels containing roads, buildings, water etc.  

\subsection{Carbon sequestration}

We assume the carbon stored is equivalent to about 50\% of the total biomass of a tree \cite{Thomas12}. %The tree biomass, above and below ground biomass, 
The AGB can be calculated for each tree based on crown diameter and its corresponding height estimation. Based on our discussion in Sec.~\ref{sec:related}, the carbon sequestered in trees approximately follows $0.65 \times \mathrm{AGB}$.
% is considered half of the total biomass.  

\section{Experiments and Results}

\begin{figure}[t]
% \vskip 0.2in
\begin{center}
\centerline{\includegraphics[width=\columnwidth]{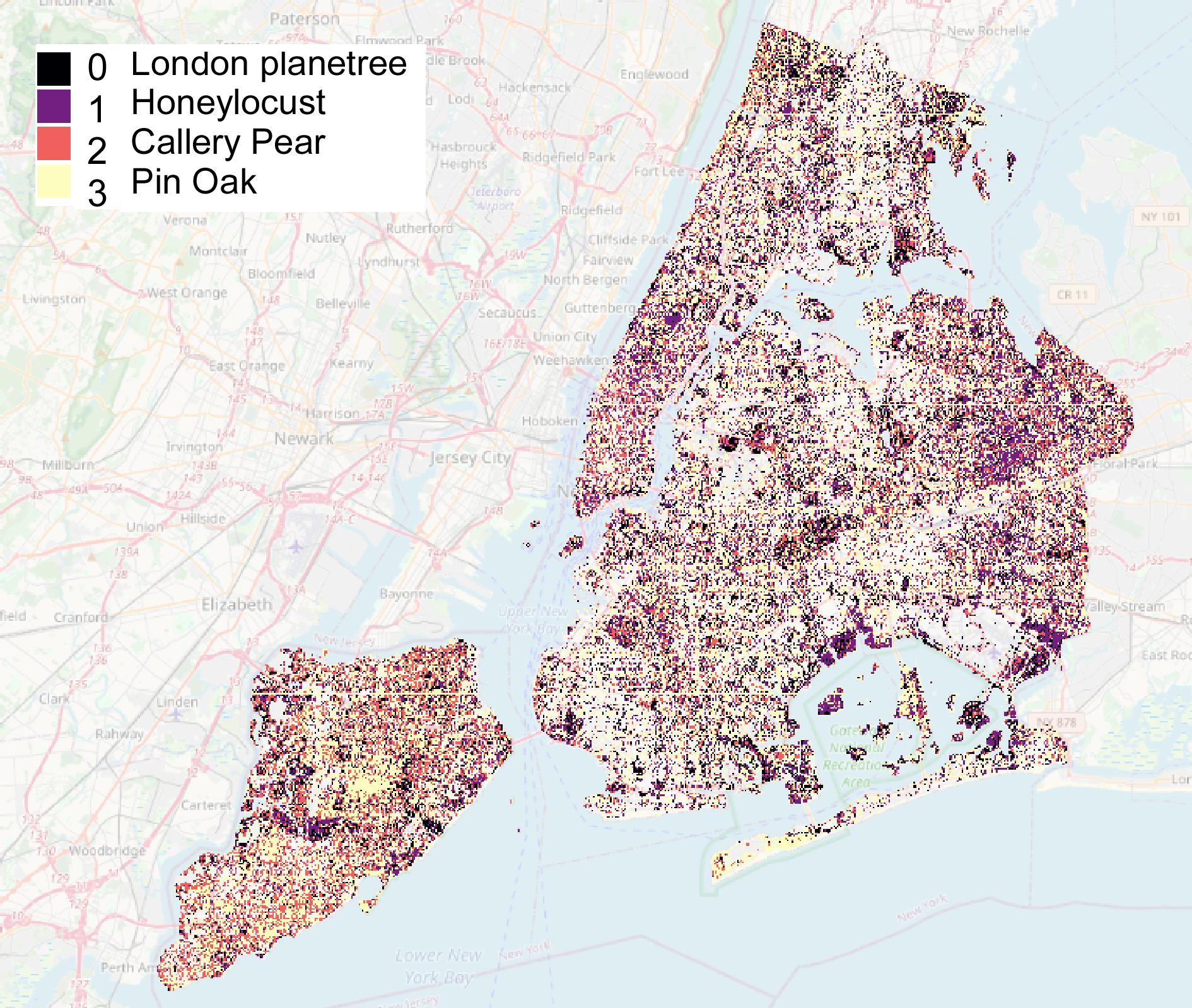}}
\caption{Tree species classification map for four tree species in New York City. The corresponding labels are referenced in Table~\ref{tab:tree_data}.}
\label{fig:species}
\end{center}
% \vskip 0.1in
\end{figure}

We apply our pipeline to the New York City (NYC) area to demonstrate our process of carbon sequestration quantification.
LiDAR data was acquired in 2017 over Staten Island borough, NYC \cite{NYCLidar}, and tree species data was collected across all five boroughs in NYC in 2015 along the public roads, only \cite{NYC15}. Trees not close to roads did not get a record in the survey---including those in parks and private properties. We processed a total of 56 NAIP tiles 50 km by 50 km in size, a total of about 50 GB in data volume. A sample view of the RGB composite depicts Fig.~\ref{fig:fitting}a.

Concerning tree delineation, an RF classifier is used to discriminate two classes---tree vs.\ anything else. Training data is manually labeled and gets employed to train the RF model that generates a tree coverage map. Once the tree-covered area is separated from the other land, individual trees is delineated by means of the watershed segmentation method. Fig.~~\ref{fig:fitting}b serves as illustration. %and tree crown diameter are estimated. %The tree crown diameters are polygonized, and polygons associated with tree crown are stored as vector objects along with other attributes like polygon area and diameter. 
Finally, the tree crown diameter is calculated based on the assumption that the crown renders approximately circular in shape, and that the diameter is proportional to the square root of the crown polygon's area.

\begin{table}[tb]
    \centering
    \caption{Data set of tree species and tree dry mass density.}
    \vskip 0.1in
    \label{tab:tree_data}
    \scalebox{1.0}{
    \begin{tabular}{l c r r}
         \hline
         Tree type & Label &  \# points &  $\rho$ (kg/m$^3$) \\
         \hline
		London plane tree    & 0 & 55,903 & 560\\
		Honeylocust          & 1 & 43,974 & 755\\
		Callery pear       & 2 & 42,384 & 690\\
		Pin oak             & 3 & 30,575 & 705\\
		
         \hline
         Total              & & 172,876 &\\
         \hline
    \end{tabular}
    }
\end{table}

We reuse the same NAIP tiles %that are  used to delineate tree covered regions 
for tree species classification. The dominant four species in NYC are sampled to generate the training data, as listed in Tab.~\ref{tab:tree_data}. The model achieves a classification accuracy of 80\% on the test split. The model is then applied across all the NAIP data after splitting into $32\times32$ tiles, cf.  \cite{autogeo} for details. The ones with a mean NDVI value lower than zero get discarded\footnote{typically non-vegetation areas like buildings, roads, water body, etc.}. A sample tree species classification map is shown in Fig.~\ref{fig:species}. The tree species of each tree's crown polygon derives from pixel-based majority vote.
% Bounding boxes of 32 by 32 pixels are classified and assigned automatically a label for the 4 tree species classes. We note that the $5\textsuperscript{th}$ class contains all other areas like  roads, buildings and water bodies. 
% For each tree crown polygon the corresponding tree species raster layer is queried and the majority returned tree species pixels are used to assign a tree species class to that tree crown polygon. In most cases, a single tree species class can be assigned to a polygon. In a smaller set of cases, two tree species classes are likely probable, and then the neighbour polygons are queries and the most dominant species is assigned to that particular polygon. 

In the next step we estimate tree heights from crown diameters. We first process the LiDAR point cloud to generate a canopy height model resulting in a height-from-ground map. For each crown polygon, we query multiple points against the LiDAR height map in order to define the corresponding mean as ground truth. 
Then, we derive a linear regression of the tree crown diameter vs.\ the tree height for each tree species. This process assembles a training data set from the LiDAR-covered areas. The linear regression curves (cf.\  Fig.~\ref{fig:fitting}c) infer the tree height from the crown polygons. % for trees identified in areas, where no Lidar data exist, and only tree species and tree crown diameters are known.

The above-ground biomass is calculated for each tree based on crown diameter and tree height taking into account the tree-specific density as listed in Tab.~\ref{tab:tree_data} \cite{WD}. We set the form factor for each tree species to $F=1$, and we estimate the AGB based on Eq.~\ref{eq:agb}. %The below ground biomass is assumed to be 30\% of the above ground value. 
Total biomass depends on tree sizes as depicted by Fig.~\ref{fig:fitting}d for Pin oaks.  
%We note, that carbon storage values are strongly dependent on tree species as both the tree density and geometrical sizes are species dependent.

\begin{figure}[t]
% \vskip 0.2in
\begin{center}
\centerline{\includegraphics[width=\columnwidth]{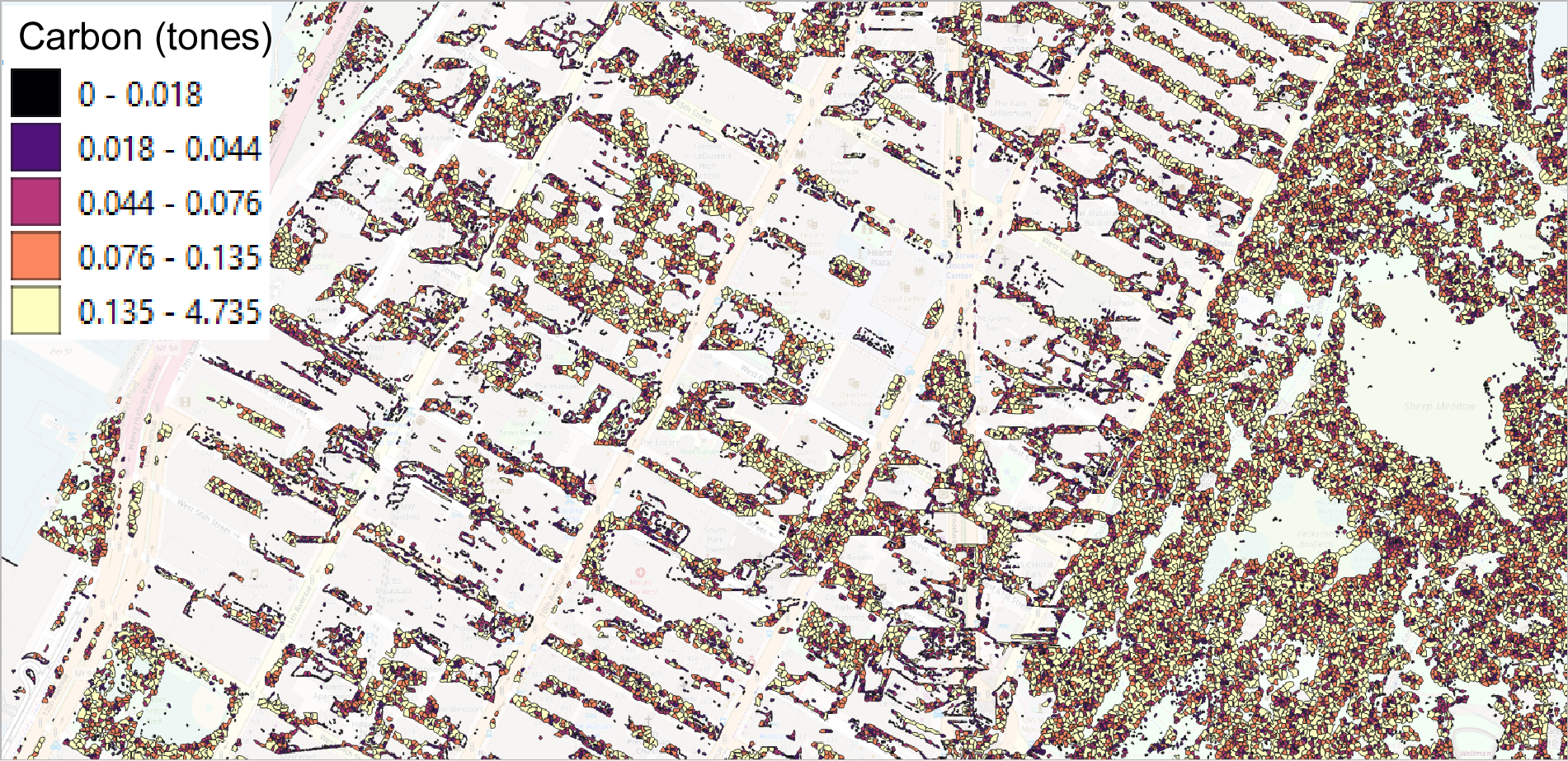}}
\caption{Carbon stored in individual trees mapped for the Upper West Side of NYC. For visual \& geographic orientation: the lower-left corner of the image hosts the Central Park.}
\label{fig:carbon}
\end{center}
% \vskip -0.1in
\end{figure}
 
A resulting map of carbon sequestered in trees is depicted by Fig.~\ref{fig:carbon} for the area of the Upper West Side in Manhattan, NYC.  %The carbon sequestration map can reproduce the spatial pattern of the carbon stored in urban forest down to individual tree level. 
We estimate the total amount of carbon stored in the urban forests of Manhattan to about $52,000$ tons. We base this calculation on summing up the results of each individual tree segmented in the NAIP data.  \cite{NYCUF18} approximates 1.2 million tons for the total carbon stored in the whole of NYC's urban forests. Breaking down this figure in proportion to the ratio of the number of trees in Manhattan versus the total number of such in NYC, the carbon stored in Manhattan trees is $\sim43,500$ tons. Carbon sequestration in Manhattan neighborhood trees is quantitatively consistent with our estimation to the order of magnitude.
%The difference between the two estimates can come from non uniform distribution of trees and the overall number of tree species considered. % \textcolor{red}{One idea: would be interesting to have a plot showing the total carbon storage for each of the five boroughs? although we might not have the space to do so.. Ideally if there is some baseline that we can compare against, to verify that the results presented here are legitimate and accurate.}

\section{Conclusions}

Precise quantification of carbon sequestration on individual tree level may enable an improved carbon trading marketplace where such information is shared in aggregated figures, only. Here we demonstrated an approach to estimate carbon stored in urban forests built on public data sets. We use aerial imagery, high-quality 3D LiDAR point cloud data, and tree species surveys to build high-resolution carbon sequestration maps. The methodology allows to map carbon sequestered by individual trees for subsequent aggregation to the level of city street to continental scale. 

%\section*{Acknowledgements}

%\textbf{Do not} include acknowledgements in the initial version of
%the paper submitted for blind review.

%\clearpage

\section*{Broader Impact}

Carbon trading markets and GHG offset require transparent and verifiable methods to quantify the total carbon sequestration. The ``bottom-up'' approach introduced in this work is able to estimate total amounts of carbon sequestered in trees. As a valuable result, a temporal sequence of spatial maps indicating the carbon density captured by (sub-)urban forests may track changes in carbon sequestration on an annual basis.

\bibliography{ref}
\bibliographystyle{icml2021}

\end{document}